# Heart Diseases Prediction Using Block-chain and Machine Learning

**Muhammad Shoaib Farooq[1], Kiran Amjad[1]**

[1] Department of Artificial Intelligence, University of Management and Technology, Lahore, 54000, Pakistan
Corresponding author: Muhammad Shoaib Farooq (e-mail: shoaib.farooq@umt.edu.pk)

**ABSTRACT** Most people around the globe are dying due to heart disease. The main reason behind the rapid increase in the death rate due to heart disease is that there is no infrastructure developed for the healthcare department that can provide a secure way of data storage and transmission. Due to redundancy in the patient data, it is difficult for cardiac Professionals to predict the disease early on. This rapid increase in the death rate due to heart disease can be controlled by monitoring and eliminating some of the key attributes in the early stages such as blood pressure, cholesterol level, body weight, and addiction to smoking. Patient data can be monitored by cardiac Professionals (Cp) by using the advanced framework in the healthcare departments. Blockchain is the world's most reliable provider. The use of advanced systems in the healthcare departments providing new ways of dealing with diseases has been developed as well. In this article Machine Learning (ML) algorithm known as a sine-cosine weighted k-nearest neighbor (SCA-WKNN) is used for predicting the Hearth disease with the maximum accuracy among the existing approaches. Blockchain technology has been used in the research to secure the data throughout the session and can give more accurate results using this technology. The performance of the system can be improved by using this algorithm and the dataset proposed has been improved by using different resources as well. Numerous algorithms have been compared on basis of their accuracy, F-measure, recall, and root-mean-square value to get better results. Novelty in this research is to find a model or algorithm that can secure the patient's data using higher accuracy algorithms. Moreover, research shows that SCA-WKNN accomplishes 15.51% greatest precision than the existing approaches, separately. Lastly, Blockchain has achieved maximum throughput than any other network in the era of security.

## I. INTRODUCTION

The heart is one of the most essential organs or muscles within the human body consisting of four chambers and the main functionality of the heart is that it works as a pump to circulate the blood [1]. The circulated blood carries oxygen that reaches the organ of the human body and leaves with carbon dioxide and wastes from the metabolism that's why the heart is known as the most important organ for human survival [2].

The term Heart disease refers to serval different heart conditions but the most common type is cardiovascular disease (CVDs) with the highest death rate [3]. the major factors behind these diseases are High blood pressure High Cholesterol levels and addiction to smoking [4].

According to research, most people in the world have three of these factors that's why most of the deaths in the world are because by heart diseases [5]. This raising tally of death due to heart disease can be controlled if patients suffering from it are diagnosed at an early stage of disease by assessing or monitoring their blood pressure levels, cholesterol levels body weight, and blood sugar levels [6]. However, the routine mentoring and maintenance of each patient's record is a difficult task to achieve [7], [8].

The current systems which are being used in our health care departments are not up to the mark as these systems do not support the latest technologies. This issue can be resolved by using technology like (IoT) internet of things which is advanced technology by using IoT devices different activities can be performed in a quick time like the healthcare professionals performing the routine checkup can collect data quickly without performing tests [9].

IoT (internet of things) can also help in monitoring the patients after the checkups. It can also be used to predict the disease at an early stage from the routine mentoring of patient data by the symptoms gathered from patients' data from different IoT devices the only problem in the current scenario is to store the vast amount of data [10].

In the architecture Design of Blockchain and heart disease management system, all the devices or objects are connected through a central server that controls and manages the access as this approach is centralized. It is not suitable for processing as well as for keeping the record of all the patients confidential which is the main aim while dealing with healthcare information. This problem can be solved by using a decentralized architecture design or which is by integrating the Blockchain technology with the current healthcare system once the block is filled it is then linked to the previous block using the technique of cryptography [11], [12].

A decentralized approach like block-chain is the most suitable in the healthcare department because the block-chain technique is synchronized as all the information or record of the patients can be stored in blocks in digital form. Once the block is filled it is linked to the previous block and this data can only be accessed by the authorized person as each block has its hash values which means the information or data stored in the blocks can only be carried out by using the hash value another advantage of using Blockchain is its immutability [13], [14].

Once the record of patients is stored in the blockhead then it cannot be modified, deleted, or destroyed by any untheorized person. The records of the patients can be kept confidential without any third-party services for that it is also called Distributed Ledger Technology (DLT). Machine learning algorithms can be used to develop a model for the prediction of any disease Med-share is a medical data sharing system that is developed for providing cloud-based services and is based on a blockchain.

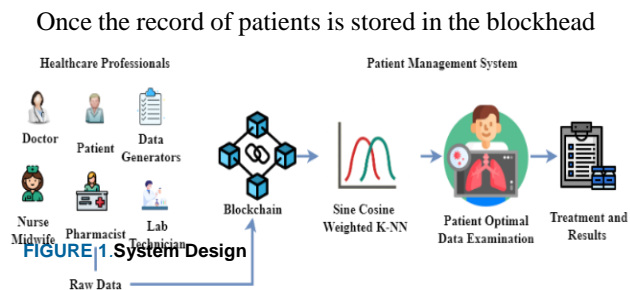

FIGURE 1. System Design

Med block is a block-chain-based medium for sharing the record or information of the patients while keeping it confidential, another framework known as MEdge-chain. It is used for sharing patient data within the management system between different entities in a secure way all these frameworks can be used to keep the information of every patient private by using a blockchain [15], [16].

The most common heart diseases are Coronary Artery Disease, Heart Valve Disease, Arrhythmia, and cardiac arrest which are also very dangerous. In this research. The heart diseases caused by blood vessel disorder will be discussed all these lives can be saved by early diagnosis of the heart diseases for which a system or model is needed which can be developed by using the machine learning algorithm like Sine-Cosine based weighted K-Nearest Neighbor (KNN) which can predict the Heart diseases [17], [18]. Figure 1 shows the complete design of the system form Healthcare staff involved in the management of the patient record, both the K-Nearest neighbor (KNN) and the Weighted KNN machine learning algorithms have their limitations and are not preferred in the current approach so, the Sine cosine algorithm in combination with weighted K-NN is used to develop a model which can predict the heart disease [19].

The system or model which has been proposed above is developed using a test network of Ethereum which is a public chain network. The model was then trained and tested using the dataset provided by the UCI repository. The system was evaluated based on performance, accuracy, precision, recall time, root mean square, and the f-score, and from the obtained results an increase of 9.95% in the accuracy of the prediction system can be seen which is higher as compared to other Machine Learning Algorithm (ML) [20], [21].

The primary goal of this research is to propose a framework to manage patients record in healthcare departments. In the past, many studies have been conducted to identify a suitable framework for heart disease prediction and managing the patient's record Currently there is no such framework has been proposed. In the research, Blockchain technology and machine learning algorithms are integrated with the current healthcare technology to develop a model that can predict and maintain the patient record effectively with maximum accuracy. Datasets from different sources are used to check the effectiveness of the proposed framework. Moreover, a Blockchain-based metaheuristics approach has also been proposed.

In this research paper, the First Section contains the introductions, as well as the significance of the Blockchain in the storing and keeping of confidential records of patients, which is also highlighted. Whereas the second Section contains the related work and the Third Section describes the methodology. The fourth Section depicts the design and analysis and the layered structure of the proposed system and the algorithms used in the research. The performance of the proposed system is evaluated using different graphs and the result are concluded in the Fifth Section, finally, the conclusion will be in the Sixth Section.

## 2. RELATED WORK

In the Healthcare Department, it is very important to keep the record of the patients confidential as the healthcare data is very sensitive. The systems which are used in healthcare departments currently are unable to address major issues like scalability, data privacy, and interoperability [22]. However, In the proposed framework Blockchain technology has been used to provide a platform for secure data transmission. Currently used systems store and process the data with the involvement of a third party [23], However, healthcare departments can use Blockchain technology as it enables the direct transaction.

The Patient data can be stored in blocks that are interlinked with each other the stored data will be distributed across the Blockchain network (BCN) [24]. However, the proposed machine learning algorithm will be applied to the patient's data for heart disease prediction.

The healthcare departments are using a system that is not integrated with blockchain technology, machine learning algorithms are also not being used in the healthcare department. Currently, healthcare departments are facing a major issue of data privacy [25], [26]. However, Blockchain Network Provides a secure key generated after the distribution of data in the Blockchain network. and the owner of the data can use the key to access or decrypt the data. Blockchain provides a secure way of sharing information between patients and cardiac professionals (CP) without compromising privacy. Blockchain is highly significant in addressing the major issue in the healthcare department such as scalability, data privacy, and interoperability.

In the past, there is no use of machine learning algorithms or any framework that can be used to keep the record of the healthcare department secure. However, in the proposed research SHA-256 and Rivest-Shamir-Adleman algorithms were used to keep the patients' uploaded records secure. There are several different methodologies and frameworks which can be used in healthcare departments for managing the medical record. These methods lack efficiency and are not cost-effective However, the proposed research aims to initiate a system that can provide a platform for the healthcare departments where the information can be shared securely and can provide data privacy. In the health care department, there is a continuous sharing of information so the chances of attacks are very high [27]. However, these attacks can be prevented by using a Blockchain network(BCN).

A set of different features have been selected to obtain the maximum accuracy and for the identification of the suitable attributes which can help in predicting the right disease [28], [29]. Moreover, a nature-inspired metaheuristic algorithm is used which has the capability of selecting a set of features that can enhance the classifier as well as the accuracy of the model. For the identification of patients with similar symptoms, an algorithm known as Particle Swarm Optimization a combination win k-means was used whereas the disease was predicted based on the distance between clusters for the prediction of cardiovascular but the accuracy can be enhanced using other algorithms.

Currently, there is no single infrastructure developed for the healthcare department that can provide secure data storage and sharing. In past research, a cloud of things (COT) technology refers to IoT solutions to deliver cloud services to a single vendor. Cloudlets and Ad-hoc are extensions of centralized cloud services and do not provide any secure way of data transmission [30]. However, in the proposed research a decentralized infrastructure is used in which patients' data is stored in the blocks forming a chain.

A patient's data chain becomes immutable and is secure even if one or two nodes are compromised as the network remains consistent. Alexandra Cernan et al. [31] .in his research proposed personal healthcare (PHR) blockchain-based system in which specific sensors were used to collect information from the patient medical records. The proposed system provides all the features mentioned above Moreover,

In the proposed system IoT devices, are used to monitor patients' data daily on which Machine learning algorithms are applied for the early prediction of heart disease in a patient. A decision tree algorithm was applied to a clinical dataset for the detection of diabetes in research conducted by Amin ul Haq et al. [32]. on the other hand, in the proposed framework an SCA-WKNN machine learning algorithm was used for the prediction of heart disease as it provided 15.51% greatest precision as compared to other classifiers.

In the research of Eman M. Abou-Nassar et al. [33]. Different technologies such as the DIT framework and a C # implementation using Ethereum were used However in addition to these technologies to keep the patient data secure distributed ledger technology (DLT) was used without using any third-party services. A signing algorithm was used for the encryption and decryption of the messages. Senthilkumar Mohan et al.

[34]. Their research uses a hybrid random forest with a linear model to enhance the performance level and with maximum accuracy, however, the proposed framework has the highest accuracy as compared to other approaches which can be seen in Figure 9. A heart disease prediction system (HDPS) was proposed by AH Chen et al. [35]. uses 13 important features and an artificial neural network algorithm for classification with a prediction accuracy of 80%, on the other hand, the proposed framework has a prediction accuracy of 97.01% as shown in table VI.

Milan Sai et al. [16]. in their research compared different machine learning algorithms for the prediction of heart disease and achieved the highest accuracy of 90.16% However the proposed framework has an accuracy of 97.01% as shown in table VI.

Mohammad Ayoub Khan [37]. proposed a lot-based framework using a modified deep neural network for heart disease prediction but the current approach is not reliable as it does not provide a secure way for data transmission, However, in the proposed framework blockchain-based decentralized database is used to keep the patient's data secure and has the highest heart disease predication than existing classifiers.

Yasser D.-Otaibi et al. [38]. proposed a framework in their research to keep patient data confidential. The framework has an accuracy of 95.8% and a response time of 1.5% However, In the proposed framework provides a secure way for data transmission as the data is stored in a decentralized database and the message is encrypted and decrypted by using a signing algorithm which is more reliable able than the above-mentioned approach. The proposed system has an accuracy of 97.01% and a response time of 2.15 (ms). Tsung –Ting Kuo et al. [39]. in their research highlighted the pros and cons of blockchain technology in the biomedical and healthcare domain whereas in the proposed framework machine learning algorithm and blockchain technology have been proposed.

Several different parameters such number of smoked cigarettes per day and the body mass index of the patient were used. Random Forest and liner model can be significant in finding the relevant features with an accuracy of 80% which can be further increased by using Naïve Byes and Logistics regression [40] [41]. However, the novelty of the proposed system is that it provides infrastructure for the healthcare department where the data can be transmitted securely by using blockchain technology but it also has the highest heart disease prediction accuracy of 97.01 to the existing frameworks.

In the past [42] [43]. A model developed using the Artificial Neural Network (ANN) has been prose which is 80% accurate in predicting the disease, therefore, it can be seen that Machine Learning (ML) and Deep Learning (DL) Algorithms can be used for storing and predicting the disease. Perhaps in the proposed research SCA-WKNN, SVM, KNN, and blockchain technology were used for heart disease prediction.

For the selection of the different sets of attributes genetic algorithm was used, for the prediction of heart disease suitable parameters such as coefficient of momentum were selected using a genetic algorithm all this process was carried out by using Coactive Neuro–Fuzzy Interface System (CNFIS) [44], [45]. However, a serval machine learning algorithm was used.

Table III shows a comparison between the proposed system and contributions by others related to this work. Six different Block chain algorithms which can be used for predicting heart disease has mentioned in the columns and the references of contributed work are mentioned in the rows. From the table above it can be concluded that all the Blockchain algorithms were used in the current research. Whereas the references show that none of these Blockchain algorithms were used in past for the prediction of heart disease. Whereas only three out of six Block chain algorithms were used in the research paper to predict heart disease.

**TABLE I: Comparison of Proposed system with Current Systems**

| Blockchain Algorithms | Heart Disease Management System (HDMS) | [14] | [15] | [16] | [17] | [18] |
|---|---|---|---|---|---|---|
| Consensus Algorithm | ✓ | ✗ | ✗ | ✗ | ✗ | ✓ |
| Chain Security Algorithm | ✓ | ✗ | ✗ | ✗ | ✗ | ✗ |
| Smart Contract | ✓ | ✗ | ✗ | ✗ | ✗ | ✗ |
| Proof of Work (POW) | ✓ | ✗ | ✗ | ✗ | ✗ | ✓ |
| Proof of Stack (POS) | ✓ | ✗ | ✗ | ✗ | ✗ | ✓ |
| Proof of Importance (POI) | ✓ | ✗ | ✗ | ✗ | ✓ | ✗ |
| Ripple | ✓ | ✗ | ✗ | ✗ | ✓ | ✗ |

### III. Materials and Methods

The process of data (messages) encryption and decryption is shown in figure 1. In the proposed framework singing algorithm has been used to validate the sender's id and to keep the messages secure. A set of keys is generated by using the RSA algorithm that provides fast encryption by keeping proof of owner authentication. For encryption of data public key is used whereas for decryption of data private key is used. only the authorized person with the key can access data in the blockchain.

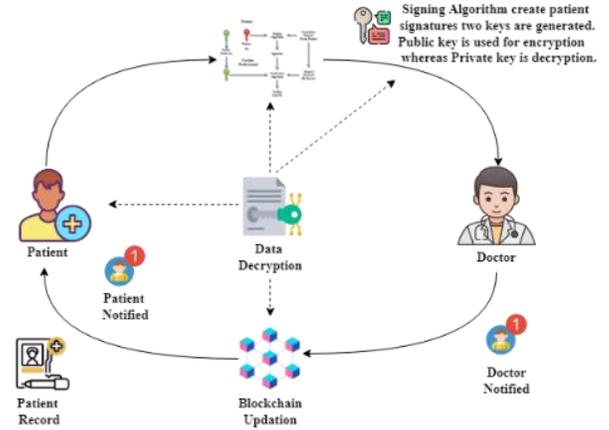

**FIGURE 2: Basic Framework of the proposed system**

Figure 3 below shows the complete architectural view of the system including the external system which is associated with the proposed system. From the architectural view of the system, both internal and external behavior of the system can be seen as all the operations performed by the system are clearly illustrated in the picture. So that it can be demonstrated how Blockchain can be used to extract, store and retrieve the data or records of the patients while keeping it confidential.

### A. MECHANISM FOR PATIENT DATA USING BLOCKCHAIN

Several Different mechanisms have been used for managing patient data using Blockchain. The patient's Data will be stored in the blocks of the Blockchain which are connected. The data of the new patient will be added to the new block. Once the data is entered a key will be generated after the data is distributed across the Blockchain network (BCN). The key can be used by the patient and cardiac professional after getting permission from the patient to analyze the record and predict the disease.

### B. MANAGING PATIENT RECORD USING BLOCKCHAIN

The patient's data is stored in form of blocks once the block is filled then the new data is stored in the new block which is then connected to the previous block. Each block has a key that is used to manage the personal data then the patient's data is distributed throughout the blockchain.

To make it assessable for the healthcare professionals which are in the system or network for managing records of patienAlgorithm1 is used in which data is collected from patient and healthcare department as input and shows how data is formed using blockchain. Machine learning algorithms have been used to select the best class label and predict heart disease in a patient.

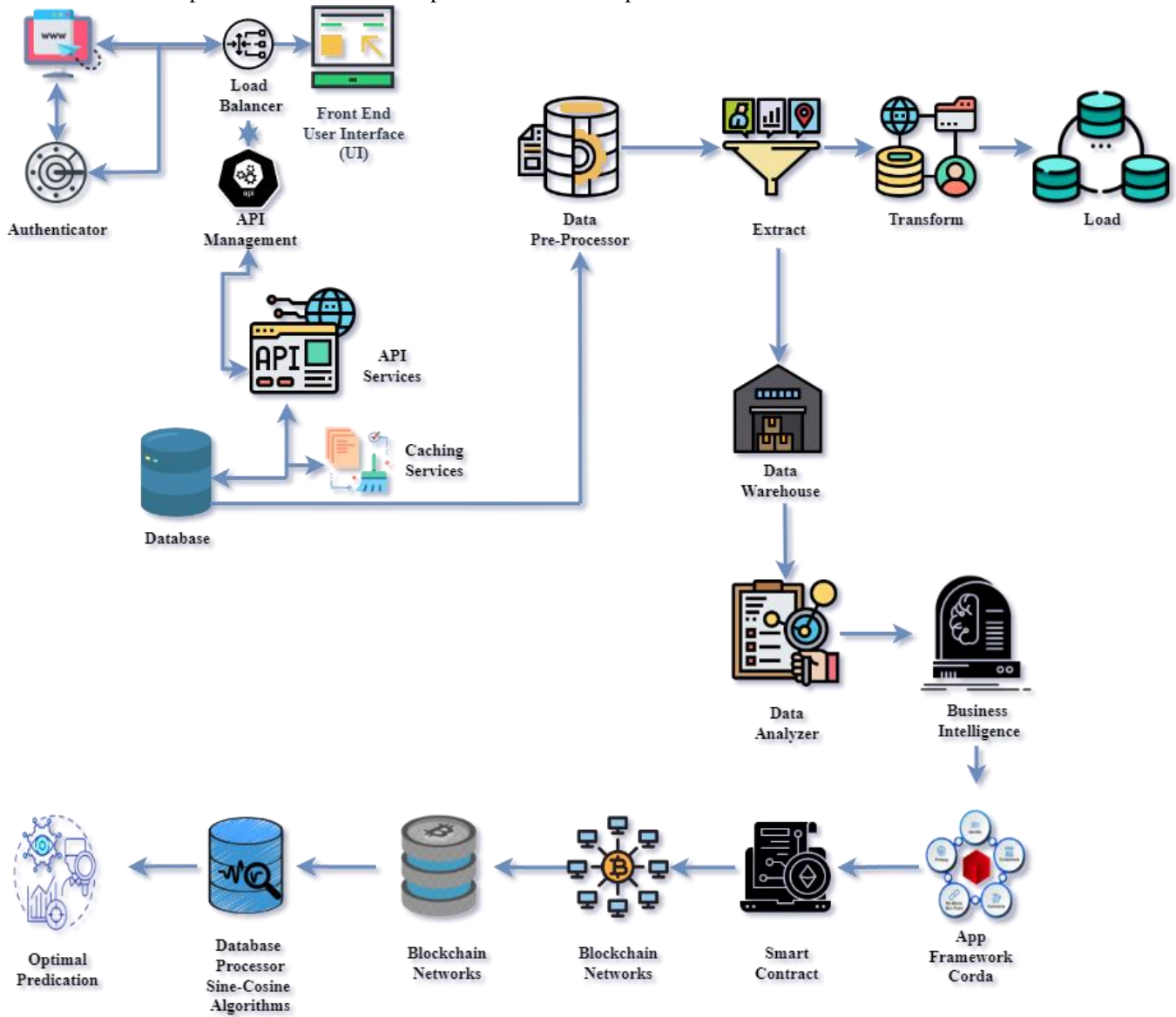

**FIGURE 3.. Architectural Design of Blockchain-Based System for Patient Record**

## C. ARCHITECTURAL VIEW OF THE SYSTEM

The architectural view of the system is illustrated in Figure 2 above from where it can be seen that the system is consisting of different components. Which includes a data processing unit that will process all the records of the patients from the extraction of the data to the retrieval of the information. Another component of the system is an application server which will be used to transfer records of the patent along with the other component the major functions which can be performed by the system are inserting the patient's record; permission to access the record, data, or record of the patient. The data is further divided into two parts to clean the data before applying the Machine Learning (ML) algorithms. Structured data will be stored in the traditional database whereas the unstructured data will be thrown into the data warehouse. In the next step, the data is sorted and then Machine Learning Algorithms are applied so that the highest accuracy can be achieved while predicting the disease. The processed data is stored in blocks using the block-chain technology, then the sine and cosine algorithm in combination with the weighted K-NN algorithm are applied for the prediction of heart disease. The collected data is retrieved from the decentralized blockchain network where it was loaded in the first place by using SCA-WKNN.

**TABLE II**: Comparison of Proposed system with Current Systems

| Notation | Description |
|---|---|
| $P_{id}$ | Patient id |
| PKPid | The private key of patient |
| $HC_{Pid}$ | Id of Health care professional |
| BCN | Block Chain Network |
| AttackerA | One who tries to steal health information |
| $Req_{HCPid}$ | Request from Health care professional |
| Acci | Represents the accuracy of the ith solution |
| $A_i$ | Represents $i_{th}$ agent |
| $A_{i,d}^{t+1}$ | Represents the value of the depth dimension of ith solution at iteration t + 1 |
| ABest | Best Solution |

## D. PATIENTS DATA PROCESSING USING WEIGHTED K-NN AND SINE COSINE ALGORITHM

As the patient's data stored in the Blockchain decentralized database is not one-dimensional. The prediction of the disease, as well as the processing of the patient's data, is not easy moreover the data of the healthcare department is very sensitive and needs more security and accuracy. The right disease can be predicated after examining the patient's records this can be achieved by using different Machine Learning (ML) algorithms.

In this research, the sine cosine algorithm is a metaheuristic algorithm that not only provides maximum accuracy but it is also the most suitable option for optimal selection due to its fast processing speed. It does not search through the whole location and can provide the desired result in less computation by finding the optimal location as compared to other algorithms. In the case of finding optimal weight to enhance the accuracy of prediction and diagnosis of disease after assessing the patient record. Each agent maps to a reliable solution that shows the weight for each k neighbor and for calculating the fitness accuracy and selected attributes can be used and it should be maximum when used with any classifier algorithm.

$$Maxf(A_i) = Acc_i \ldots \ldots (1)$$

Were,

$A_{cci}$ donates the accuracy of the ith agent

The position of the agent can be calculated using Equations (1) and (2) after every iteration.

**TABLE III. Matrices Evaluation**

| Metrics | Formula |
|---|---|
| Accuracy | $\dfrac{TP + TN}{TP + TN + FP + FN}$ |
| Precision | $\dfrac{TP}{TP + FP}$ |
| Recall | $\dfrac{TP}{TP + FN}$ |
| F-Measure | $2 * \dfrac{precision * recall}{precision * recall}$ |
| Root Mean Square Error | $\sqrt{\dfrac{1}{n} \sum_{i=1}^{n} (x_{i\_} x^{\wedge}{}_i)^2}$ |

$$A_{i,d}^{t+1} = A_{i,d}^t + x_1 * cos_{(x2)}$$
$$+ |x_3 * A_{Best,d}^t - A_{i,d}^t| If(x4$$
$$< 0.5) \dots (4)$$

Were, $A_{i,d}^{t+1}$ is for the position of i for each dimension data and iteration t

$A_{Best,d}^t$ donates the suitable agent at the iteration t

Let x1, control the exploration

$$x_{1=a-t*\frac{a}{T_,}} \dots (5)$$

t donates the current iteration

T is for the total number of iterations required for the coverage of the fitness. $\alpha$ is a constant initialized to two and goes all the way to zero where x2 checks the movement of the agents toward the best agent x3 assigns the weight and x4 is to move between sine and cosine functions.

For the prediction of the disease using the sine and cosine algorithm and for finding the optimal weights and the K-NN algorithm is shown in Algorithm 4. The Euclidean distance can be calculated by using new patient data with an object of the data set for which the K-NN with the shortest distance is chosen for SCA_WKNN at each iteration the value can be updated using Equations (2) and (3). The agent with the highest number of iterations and label class is assigned to new patient data. Algorithm 5 also shows the workflow of the SCAWKNN for the prediction of heart disease.

Figure 4 shows the behavior of the proposed system. The Health disease management system (HDMS) will be managed by an administrator. The administrator will manage all the activities of the proposed system by creating an account. to perform different tasks, the administrator will log in to the system by using his/her Id and password the system will verify the credentials of the administrator in case of authentic credentials the administrator will be allowed to login into the system. The administrator will reenter the Id and Password in case e of invalid authentication. The administrator will be allowed to perform a transaction and select a patient and the data of the patient will be added to the block and will be uploaded to the Blockchain network and a digital signature will be added. The transaction will only be validated and allowed if the provided digital signatures are valid otherwise the system will exit and the transaction will not be allowed.

In the Blockchain layer after collecting the data from the user public and private Blockchain networks can be created once the data is distributed across the blockchain network. To access the data, the users with the key can access on other hand the unauthorized user can only access the data after getting permission from the authorized person with the key and the record of the user can be accessed. To predict heart disease this process is shown in the transaction layer and the last layer show how a database which is a decentralized database in this case because a decentralized database is more secure than a centralized database.

In Blockchain decentralized database once the data is stored and distributed among the nodes than the record than the owner of the data can only decrypt the data by using a key as the Blockchain is also secured by consensus algorithms such as proof of work (POW) or Proof of stake (POS)., this task can be performed by the user anonymously without affecting the transparency of the record.
The methodology of the system is defined in different layers or stages of the layered Structure diagram which is shown in figure 3. The workflow of the system is divided into six different layers and each layer has its functionality.

### 1) LAYER 1: DATA COLLECTION
Patient data will be collected via HDMS (DAPP) the collected data will be distributed to the blockchain network. To access patient data, the cardia professional needs to request the owner for the key to analyze the patient record.

### 2). LAYER 2: NETWORK LAYER
A decentralized database is used because a decentralized database is more secure than a centralized database. In a blockchain decentralized database once the data is stored and distributed among the nodes. The owner of the data can only decrypt the data by using a key as the blockchain is also secured by consensus algorithms such as proof of work (POW) or Proof of stake (POS).

### 3) LAYER 3: TRANSACTION LAYER

The authorized person with the key can access the data whereas the unauthorized user can only access the data after getting permission from the authorized person with the key. After getting the key patients record can be accessed to analyze and predict heart disease.

### 4) LAYER 4: AUTHENTICATION LAYER

In the authentication layer, different levels of security are shown where the users have to provide certain information and a smart contract because the data in the blockchain is stored in digital ledgers. The user can also insert the data new block using blockchain algorithms such as proof of work (POW).

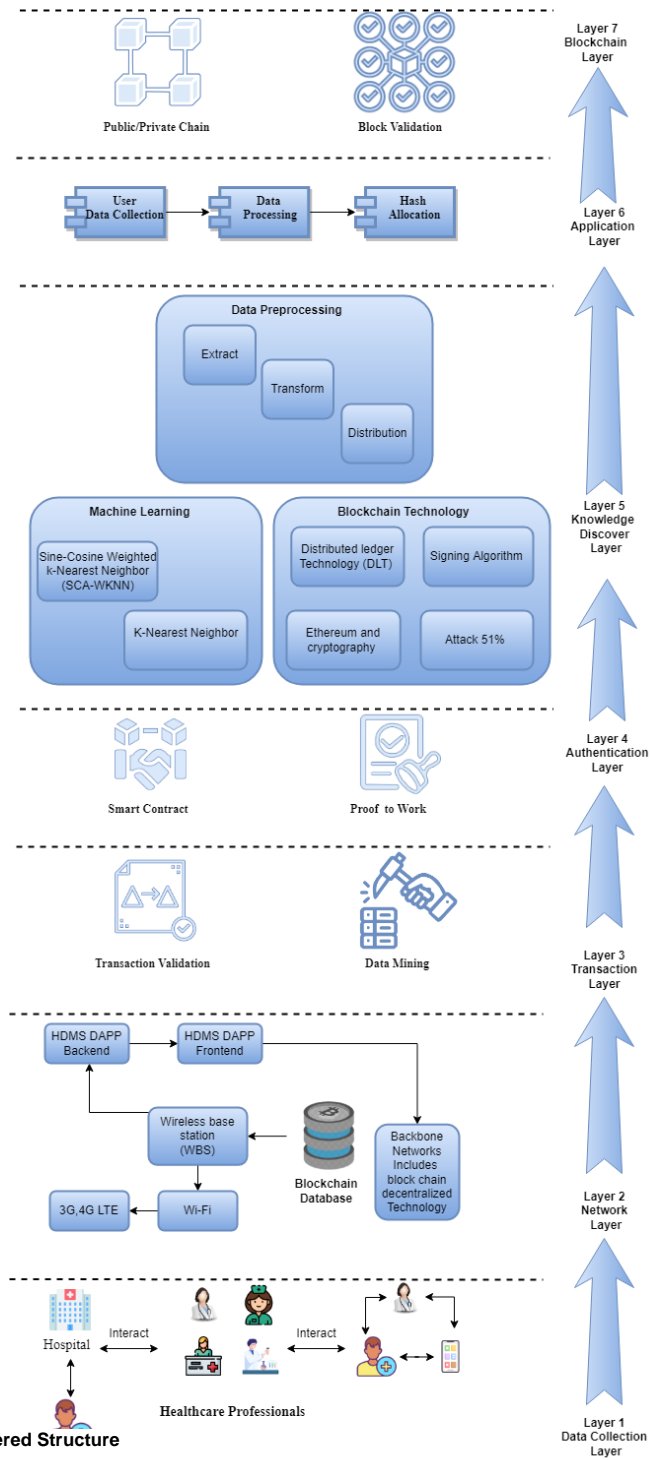

**FIGURE 4.** Layered Structure

### 5) LAYER 5: KNOWLEDGE DISCOVERY LAYER

Patient data will be preprocessed; the data will be distributed all over the blockchain network. Machine learning algorithms sine cosine weighted k-nearest neighbor (SCA_WKNN) and simple k-nearest neighbor (KNN) are applied to the patient data for heart disease prediction.

### 6) LAYER 6: APPLICATION LAYER

In the data collection patient, data will be collected. The collected data will be preprocessed after data preprocessing data will be distributed all over the blockchain network. Singing algorithms have been used for data decryption and encryption. Different keys are generated by using the RSA algorithm for this process.

### 7) LAYER 7: BLOCKCHAIN LAYER

In the Blockchain layer after collecting the data from the user. Public and private blockchain networks can be created once the data is distributed across the blockchain network.

The activity diagram Figure 4 shows the behavior of the proposed system. The Health disease management system (HDMS) will be managed by an administrator. The administrator will manage all the activities of the proposed system by creating an account. to perform different tasks, the administrator will log in to the system by using his/her Id and password the system will verify the credentials of the administrator in case of authentic credentials the administrator will be allowed to login into the system and the administrator will reenter the Id and Password in case e of invalid authentication. The administrator will be allowed to perform a transaction and select a patient and the data of the patient will be added to the block and will be uploaded to the Blockchain network and a digital signature will be added. The transaction will only be validated and allowed if the provided digital signatures are valid otherwise the system will exit and the transaction will not be allowed. A patient data chain is formed in this process in which the data is stored in the block which is interlinked with each other. In the proposed approach the data is stored in a blockchain decentralized database which is more secure as the data stored in the block will not compromise in case of any malicious attack because the network will remain consistent and the only way to access the data is by using the key as the messages are secured by using a signing algorithm.

**TABLE IV: Transactions in Heart Disease Management system (HDMS)**

| TxHASH | Block | From | To | Value |
|--------|-------|------|-----|-------|
| a6697... | 1 | 8b536. | T1SC | Patient Information (Nonce, Timestamp) Patient 01 Cardiac Professional 01 |
| 686b7... | **2** | **e7888...** | **T2SC** | **Patient Information (Nonce, Timestamp) Patient 02 Cardiac Professional 02** |
| 9ef7f.... | 3 | f451a... | T3SC | Patient Information (Nonce, Timestamp) Patient 03 Cardiac Professional 03 |

In a Blockchain network (BCN) while executing a transaction smart contracts are used to provide a secure connection to the user it is compulsory to follow the smart contract to save the data successfully. In the proposed system the smart contract is between the patient and the cardiac Professionals. Table 2 contains the Tx Hash which shows the transaction ID of the transaction, the Block column contains the block number where the transaction is being performed, and from contains the hash value of the block where the truncation is being sent.

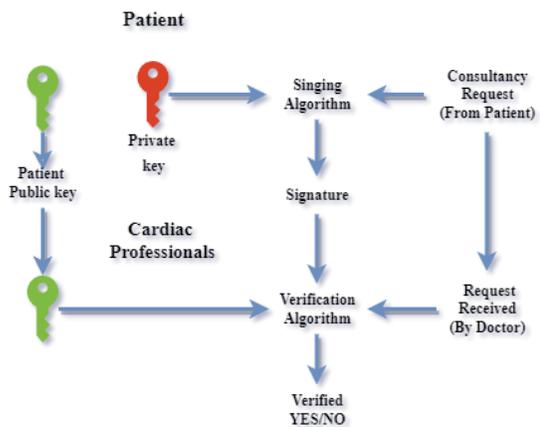

**FIGURE 5. Signing Algorithm**

The main reason for using Blockchain technology in the healthcare department is to provide a platform where the data can be shared in a secure way between the patients and the cardiac professionals (CP). The proposed system uses different keys for the encryption and decryption of the data stored in the Blockchain Network (BCN). The patient can only access or decrypt the data by using the key. The private key and message of the patient are used for creating the signatures. The private key and the message of the patient are used as parameters by a signing algorithm to create patient signatures which are then stored in the Blockchain network (BCN) along with the public key same Secret key is used for both creating and verifying the signatures. The process is shown in Figure 5.

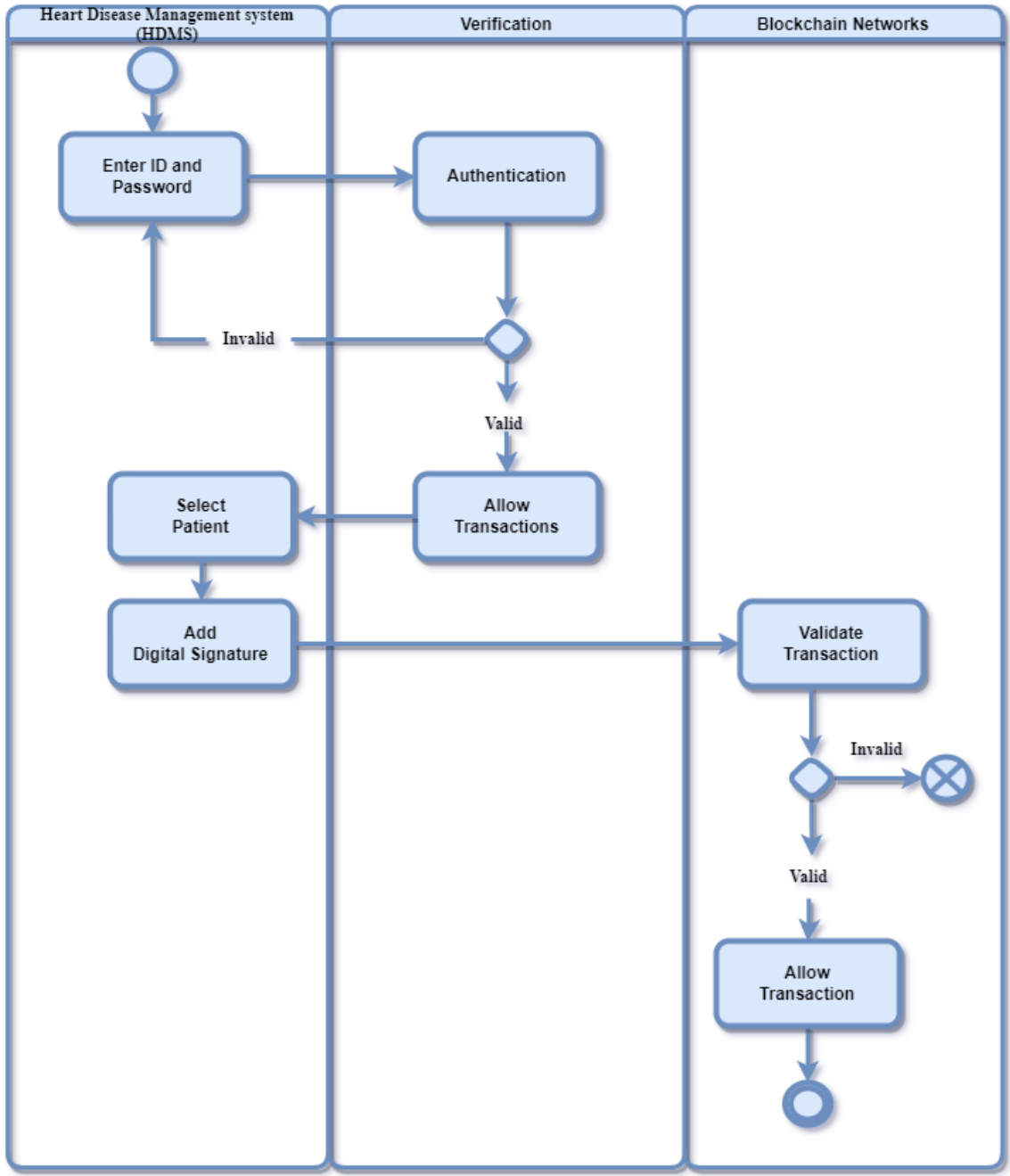

**FIGURE 6. Activity Diagram of the Proposed system**

In the Blockchain network, cryptography hashes are used to keep the data secure from the attacker. The data of the patient is encrypted by these cryptographic hash values both at the time of the transaction or when the data is added to the blocks of the Blockchain Network(BCN) only the authorized owner can decrypt the transaction using a private key. As all the data of the patient is encrypted by a has value only the nonce value can be changed by the miner. A simple block of the Blockchain network consists of a nonce value which is a random value that is generated every time as it is the number that is only used once. The data part contains the data of the patient's Previous hash and stores the hash value of the previous block whereas the hash contains the hash value of the current block. the blocks in the Blockchain network are linked with each other as shown in Figure7.

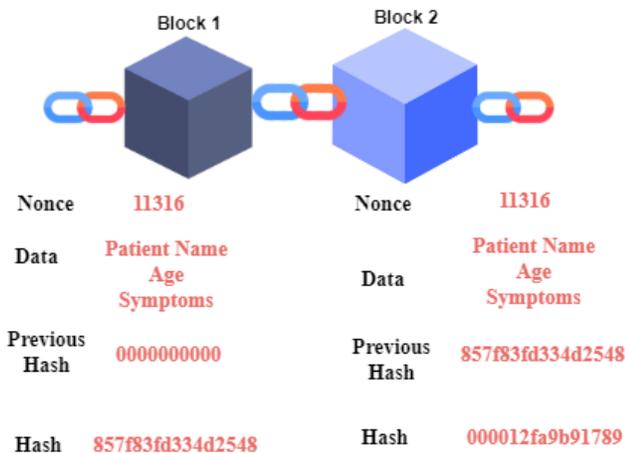

**FIGURE 7.** Blocks in Blockchain Network

The 51% attack means controlling the 51% hash rate of the Blockchain network (BCN). To control the 51% hash rate of a system with a very high computational power is required. The data of the patient is stored in the blocks of the Blockchain network (BCN).

The blocks are connected as each block contains the hash value of the previous block forming a chain of blocks. The patient record is distributed across the network and a key is generated. The patient's data is distributed in the Blockchain network so that it is visible to all the actors in the Blockchain network (BCN).

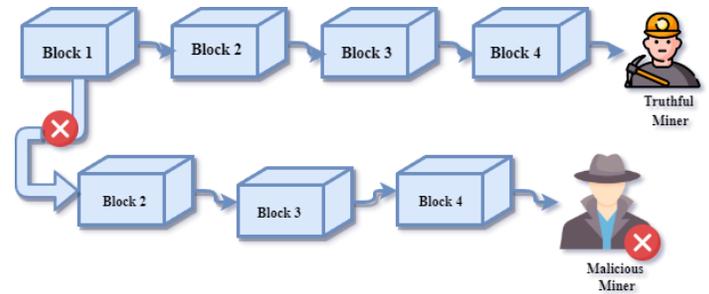

**FIGURE 8.** Attack 51 % Prevention in Hearth Disease Management system

The patient data will only be accessed and decrypted by the authorized person with a key. In the Proposed system, the cardiac professionals (CP) are the pre-s elected miners of the patient data because their cardiac professional id is verified by the user then the key is provided to access and analyze the data. To keep the process fully secure in case any authorized person tries to perform any malicious activity attack 51% will be declared and access to the particular block will not be granted as shown in Figure 8.

### E. DESIGN AND ANALYSIS
This section gives a brief overview of the design and the layered structure of the proposed system and the Blockchain network algorithm which were used during the research.

1)HEART DISEASE MANAGEMENT SYSTEM (DBAPP)

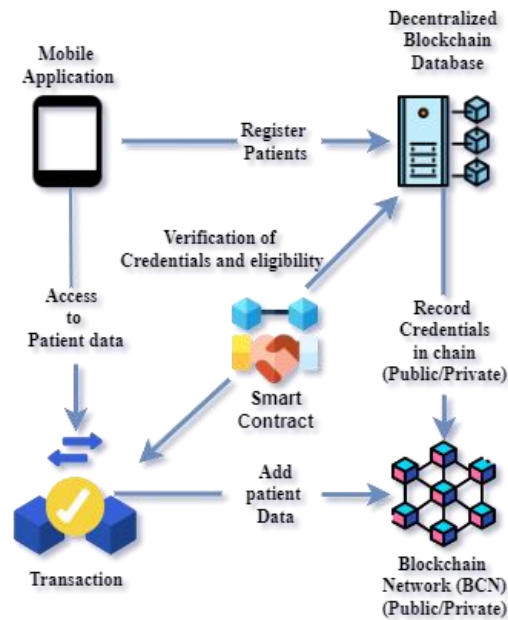

FIGURE 9. Heart Disease Management System (HDMS dbApp)

The heart disease management system consists of different components which are shown in Figure VII. The HDMS(dbApp) has a user-friendly interface and provides a secure interaction between the user and the system. The HDMS (dbapp) is developed based on Blockchain technology so the HDMS (dbapp) is a decentralized application.

The user of the application will have to register first and a unique user ID and password will be allotted to the user. The user login to the system after entering his/her credentials the system will verify the user's credentials and the user will be registered. The system provides a smart contract between the Cardiac professional (CP) and the patient then the cardiac professional (CP) can get access to the patient's data after his/her permission and can perform a transaction. After the transaction is performed by the Cardiac professional (CP) the patient data will be distrusted to the Blockchain networks and will be stored in the database.

Algorithm 1 below shows how the data of the patient and cardiac professionals is organized and processed in a Blockchain network. Both the patient Id (PID) and the cardiac professional id (CPID) are taken as input after which the data is stored in the blocks and then a systematic and private key is generated. The patient id (PID) is then distributed across the Blockchain network (BCN) so that it can be accessible by the cardiac professional (CP) and all other actors in the network in case a patient's ID (PID) is accessible by an attacker it will be removed from the network.

**Algorithm 1**

Patient Information (Patient ID $P_{id}$, Cardiac Professional $Cp_{id}$)

| | |
|---|---|
| **Input:** | Patient ID $P_{id}$, Cardiac Professionals (CP $_{id}$) |
| **Output:** | Access to patient Data PD |
| | **Begin** |
| | **For** $\forall P_{id} \in P$ |
| | Include in Block-chain BCN $\leftarrow$ BCN U {P-{id}}| |
| | Give Access ($P_{id}$, PK $_{Pid}$) |
| | **End For** |
| | **For** $\forall CP_{id} \in CP$ |
| | Include $CP_{id}$ in Block-chain BCN $\leftarrow$ BCN U {Cp$_{id}$} |
| | Give Access ($CP_{id}$, PK$_{CPid}$) |
| | **End For** |
| | **For** $\forall P_{id} \in P$ |
| | **If** Pid $\in$ AttackerA |
| | Remove $P_{id}$ from Block-chain BCN $\leftarrow$ BCN – {$P_{id}$} |
| | **End If** |
| | **End For** |
| | **For** $\forall CP_{id} \in P$ |
| | **If** $CP_{id} \in$ AttackerA |
| | Remove $CP_{id}$ from Block-chain BCN $\leftarrow$ BCN – {$P_{id}$} |
| | **End If** |
| | **End For** |
| | **End** |

The operations which can be performed by the patient in the Blockchain network (BCN) are described in Algorithm 2 below. Any Patient in the Blockchain can inset their record and is allowed to perform a transaction. The data is distributed in the decentralized based Blockchain network (BCN) patient ids (PIDs) will be updated, updated patient ids (PIDs) are visible to every actor in the Blockchain network(BCN) But these Patient Ids (PIDs) can only be accessible once permission is granted by the respective patient.

**Algorithm 2**

Operations on Patient Data ($P_{id}$, PK$_{id}$, Req $_{CPid}$)

| | |
|---|---|
| **Input:** | Patient ID $P_{id}$, Private Key of patient PK$_{pid}$, CP Request Req$_{CPid}$ |
| **Output:** | Access to patient Data PD |
| | **If** $P_{id} \in$ BCN |
| | **If** PD $P_{id} \notin$ BCN |
| | CreatePatientData ($P_{id}$, PDPid, BCN) |
| | Else |
| | UpdatePatientData ($P_{id}$, PDpid, BCN) |
| | **End If** |
| | **Else** |
| | Print Invalid $P_{id}$ |
| | **End If** |
| | **If** ReqCP$_{id}$ is Received |
| | **If** CP$_{id} \notin$ AttackerA |
| | Give Access ($CP_{id}$) |
| | **End If** |
| | **End If** |
| | **End** |

The cardiac professionals (CP) involved in the Blockchain network (BCN) can perform different operations These operations are shown below in Algorithm 3. The patient will insert their data along with the symptoms that they are experiencing in the block of the Blockchain network and the data of the patients will plod in the Blockchain. For the consultation the patient has to send a request to their desired consultant then the doctor will receive and accept the patient's request depending upon the availability, to access the patient's record the doctor will send a request to the respective patient which will be accepted if cardiac professional ID (PID) is valid and then the access will be granted to the doctor so that the doctor can analyze the patient record and predict the disease.

**Algorithm 3**

Cardiac Professionals Operations (CP$_{id}$)

| | |
|---|---|
| **Input:** | Cardiac Professional ID (CP$_{id}$) |
| **Output:** | Predict the Disease for Patient Data PD |
| | **Begin** |
| | **If** Req $_{Pid}$ is Received |
| | **If** P$_{id}$ ∉ AttackerA |
| | View Patient Data (CP$_{id}$, PD$_{Pid,}$ BCN) |
| | K_NN_Disease Predication () |
| | **End If** |
| | **End If** |
| | **End** |

Algorithm 4 above shows the significance of the sine and cosine algorithm in finding the optimal weights as well as in predicting the disease using the K-Nearest Neighbors (K-NN) Machine Learning Algorithm. In the algorithm above K-Nearest Neighbor (K-NN) is used in the beginning to find the K-Neighbor which will invoke Algorithm 5.

**Algorithm 4**

Disease Prediction Using K-NN (CP$_{id}$)

| | |
|---|---|
| **Input:** | Dataset and Patient DataPD |
| **Output:** | K Neighbours |
| | **Begin** |
| | **For** each instance $X_i$ ∉ Dataset |
| | Calculate the Euclidean distance as (PD, X$_i$) |
| | **End For** |
| | $\varphi$ = Find The K-Neighbors of PD |
| | Invoke SCA_WKNN($\varphi$) |
| | **End** |

 The framework for using the Machine Learning (ML) algorithm for predicting heart disease is shown below in Algorithm 5. Algorithm 5 shows the assignment of the class label to the instance. A simple K-Nearest Neighbors K-NN is used to find the class label of the patient on basis of the Nearest Neighbor, the Euclidean distance is calculated among different patients and the instance from the proposed dataset. The K-Neighbor with the minimum value is selected as an input for the sine cosine weighted k-nearest neighbor (SCA-WKNN). SCA-WKNN aims to find the optimal class label for the patient data. The distance between the patient data and its neighbor can be used to find a class label for different patients by assigning weights Equations (2) and (3) are used for updating the value of the weights after each iteration after the maximum number of iterations the object with the best accuracy is calculated. Every new patient is assigned a class label of the agent with the best fitness which is computed after the maximum number of iterations and the class label of that agent is assigned to the data of a new patient. The class label is key for predicting heart disease in a patient because the class label indicates if the disease is present or not in a patient. Algorithm 5 shows the working of SCAWKNN for heart disease prediction.

**Algorithm 5**

Sine Cosine Weighted K-Nearest Neighbours SCA-WKNN($\varphi$, PD)

| | |
|---|---|
| **Input:** | $\varphi$ Neighbors ,Patient Data PD |
| **Output:** | Class Label i.e. Disease Predication for PD |
| | **Begin** |
| | Initialize the agents A with $\varphi$ neighbors |
| | **While** (t!=T) |
| | **For** ∀A$_i$ |
| | Calculate fitness by using Eq (1) |
| | **End For** |
| | A$_{Best}$=(A$_i$)|$_f$(A$_i$) |
| | **For** ∀A$_i$ |
| | **If** fitness of agent A$_i$ is better than $A_{Best}$ |
| | $A_{Best}$=A$_i$ |
| | **End If** |
| | **End For** |
| | Calculate r$_1$ using Eq (4) |
| | $r_2 = GenerateRand(\$)$ |
| | $r_3 = GenerateRand(\$)$ |
| | $r_4 = GenerateRand(\$)$ |
| | **For** ∀A$_i$ |
| | **If** r$_4$ < 0.5 |
| | Calculate Position Using Eq (2) |
| | **Else** |
| | Calculate Position Using Eq (3) |
| | **End If** |
| | **End For** |
| | **End While** |
| | Assign Class Label for PD as $A_{Best}$ |
| | **End** |

## F. SMART CONTRACT

 Smart contracts are like simple programs stored on the blockchain network that is executed when the constraints applied by the user are met. Smart contracts are reliable as they automatically execute the agreements so that all the users in the contract can be certain of the results and no alteration can be made once the contract is part of the blockchain.

**Smart contract: Heart Disease Predication Smart Contract**

1. contract identity
2. string name;
3. uint age;
4. string gender;
5. string symptyoms;
6. uint id;
7. uint[] public arr;
8. uint choice;
9. string testrequired;
10. string predictresult;
11. string public result;
12. string public sugerreportresult;
13. string public cholesterolreportresult;
14. constructor() name=""; age=0; gender=""; gender="";
15. symptyoms="";
16. id=0;
17. function setname(string memory pname) public
18. name=pname;
19. function getname() view public returns(string memory)
20. return name;
21. function setage(uint page) public
22. age=page;
23. function getage() view public returns(uint)
24. return age;
25. function setgender(string memory pgender) public
26. gender=pgender;
27. function getgender() view public returns(string memory)
28. return gender;
29. function setsymptyoms(string memory psymptyoms) public
30. symptyoms=psymptyoms;
31. function getsymptyoms() view public returns(string memory)
32. return symptyoms;
33. function setid(uint pid) public
34. id=pid;
35. function getid() view public returns(uint)
36. return id;
37. function setdoctorid (uint item ) public
38. arr.push(item);
39. function setchoice(uint pchoice) public
40. choice=pchoice;
41. function getchoice() view public returns(uint,string memory)
42. string memory msg = "Cosultant is Selected";
43. return (choice,msg);
44. function settestrequired(string memory ptestrequired) public
45. testrequired=ptestrequired;
46. function gettestrequired() view public returns(string memory,, string memory)
47. string memory msg="Required Test";
48. return(testrequired,msg);
49. function setpredictresult(string memory presult) public
50. predictresult=presult;
51. function getpredictresult() view public returns(string memory)
52. return predictresult;

In the proposed research smart contract solidity and remix online compiler to develop a smart contract.

The smart contract is between the cardiac professionals and the patients. In the contract, the patient data is collected by using different functions. Different Ids will be provided to the patients, whereas cardiac professional id will be allocated to the available cardiac professionals. The patients have the privilege to select the desired cardiac professional by entering the id of the cardiac professional. The cardiac professional will examine the patient and will recommend the required test based on previous records and symptoms experienced by the respective patient. Based on the test reports and the symptoms, the cardiac professional will predict heart disease in a patient.

## IV. Experimental Results

This section provides an overview of the experimental results and a graphical analysis of the performance of the proposed system. The data processing of the proposed system has been implemented by using the combination of both Sine Cosine and k-Nearest Neighbor algorithms with the Blockchain system to check the scalability and deployment feasibility. The proposed method is then evaluated under different scenarios and actors who will interact with the system such as cardiac professionals, Doctors, and patients.

### A. DATA SET

The dataset used in the research is taken from the UCI database. There are 271 different objects and 76 critical attributes in the dataset out of which 14 most essential attributes which can lead to heart disease were selected. These key attributes include body weight, blood sugar level, cholesterol level, and pain in the upper body. The experiment aimed to analyze if a particular patient had suffered from heart disease.

### B. PERFORMANCE EVALUATION

Several different experiments were performed to calculate the efficiency of the proposed system using sine cosine weighted K-Nearest Neighbor (SCA-WKNN). The data of the healthcare department is very sensitive and need a special system that can protect the data of the patients in the Blockchain network (BCN). Metaheuristic approach was used to assign the different class label for the new object to get maximum accuracy while predicting the disease.

The algorithms of Machine Learning (ML) such as support machine vector (SVM), Random forest (RF), K-Nearest Neighbor (KNN), and weighted K-Nearest Neighbor (WKNN) were compared with the proposed sine cosine weighted K-Nearest Neighbor (SCA-WKNN) Algorithm.

The required features from these algorithms were selected after comparing them. The data set was divided into two parts on the ratio of 70:30 the first 70% was used for training and the remaining 30% was used for testing. The mean value for analysis was calculated after evaluating the model multiple times.

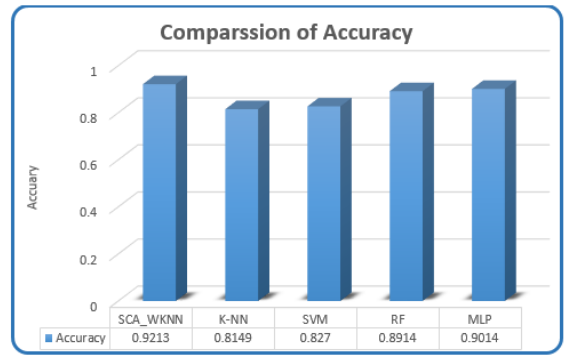

**FIGURE 11. Comparison of Accuracy**

From Figure 11 above the dataset provided in Figure 11, it is evident that the Sine cosine weighted K-nearest algorithm(SCA-WKNN) used in this research is more accurate in predicting the disease than the machine learning (ML) Algorithms.

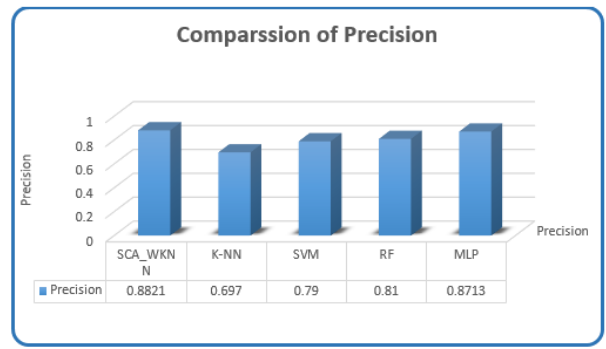

*Figure12.* Comparison of Precision

The graphical analysis of precision between different Machine Learning(ML) algorithms and the sine cosine weighted k-nearest neighbor algorithm (SCA-KNN) and the dataset is provided in Figure 10. Among all the algorithms of machine learning Sine cosine weighted K-nearest neighbor (SCA_KNN) can be used to get a high precision rate as it has the highest value of precision as shown in Figure 12.

## *C. RESULTS AND DISCUSSION*

The results obtained from the research are concluded and discussed in the following section.

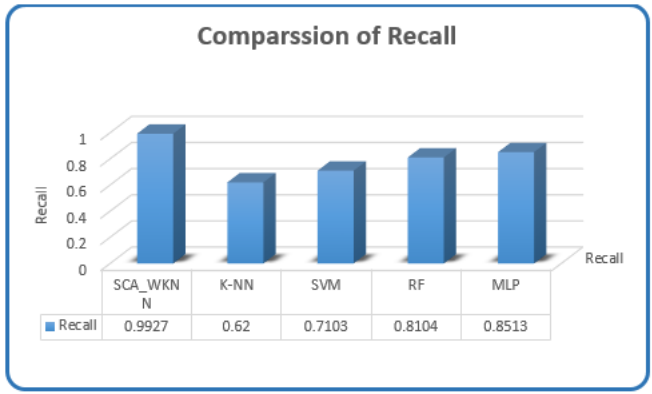

**FIGURE 13. Comparison of Recall Between Classifiers**

From Figure13 it is evident that sine-cosine weighted K-nearest Neighbor has the highest value of recall among all the classifiers.

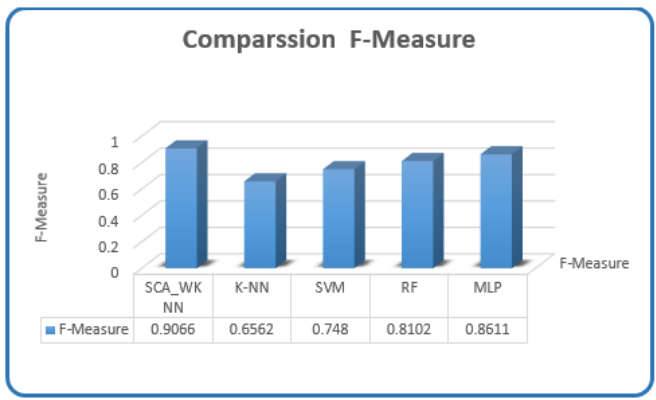

**FIGURE 14. Comparison of F-measures**

Sine cosine Weighted K-nearest Neighbor (SCA-KNN) has the highest f-measure value among all the classifiers which is shown in Figure 13.

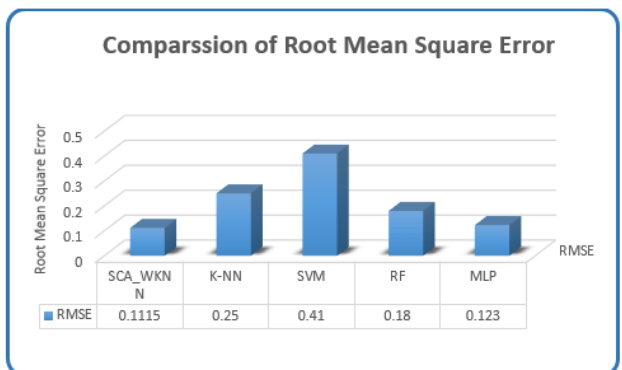

**FIGURE 15. Comparison of Root Mean**

From Figure 13 it is evident that the sine cosine weighted K-nearest neighbor (SCA-KNN) has the lowest root mean square value whereas the Support vector machine (SVM) has the highest Root mean square value.

The Blockchain data is analyzed for the prediction of the disease. From the graph above it is evident that the sine cosine weighted k-nearest neighbor has the maximum accuracy among all the classifiers where the value of k=1. The Comparison of the accuracy of several different classifiers is shown in Figure 9. The data of the health care department is very sensitive and there is a need for a highly secure and accurate system. To achieve the maximum accuracy Sine –cosine weighted K-nearest Neighbor (SCA-WKNN) can be used which is shown in Figure 9.

**TABLE V. Durations by Classifiers**

| Classifier | Time Taken (ms) |
|---|---|
| Sine Cosine Weighted K-Nearest Neighbor (SCA_WKNN) | 2.15 |
| Weighted K-Nearest Neighbors (W K-NN) | 1.835 |
| K-Nearest Neighbor (K-NN) | 1.643 |

The duration of several different algorithms is shown in Table 5. Sine cosine Weighted K-nearest neighbor (SCA-WKNN) has taken maximum time as compared to Weighted K-Nearest neighbor (WKNN) and K-Nearest Neighbor (K-NN). To find the neighbor in the data set and to find the distance between the test instance and all other instances K-nearest Neighbor is used. Both Weighted K-nearest Neighbor (WKNN) and K-Nearest Neighbor (K-NN) Require maximum time to find a neighbor for the test instance.

**TABLE VI. Compression of current and proposed Approaches**

| Methodology | Classifier | Accuracy |
|---|---|---|
|  |  |  |
| Ali F et al | Deep Learning | **85.5** |
| Fitriyani NL et al | Density-Based Spatial Clustering | **96.8** |
| Ayon SI et al | Deep Neural Network | **98.55** |
| Proposed Approach | Proposed Approach | **97.1** |

The compression of predicting heart disease by using the current approach and the existing approaches is shown in Table 6. From Table 6 it is evident that the proposed sine-cosine weighted K-nearest neighbor can achieve 15.51% greater accuracy than other approaches. The accuracy of the sine cosine weighted K-nearest neighbor (SCA-WKNN) can be increased for predicting optimal heart disease after further research.

**Table *VII*. Comparative analysis of Blockchain algorithm for a set of Blockchain properties**

| Parameters /Algorithms | Proof of Work (POW) | Proof Of Stake (POS) | Delegated Proof of stake (DPOS) |
|---|---|---|---|
| Developers | Markus Jakobson and Ari Jules | Peercoin | Danial Larimer |
| Year | 1999 | - | 2014 |
| Node Identification | Public | Public | Public |
| Computational Power | High | Partial | low |
| Efficiency of Energy | No | Low | Partial |
| Data Model | Transaction-based | Account-based | Transaction based. Account-Based |
| Application | Crypto-currency. General application | Michaelson Application | Decentralized Exchange |
| Language | C++, Golang Solidity | Michaelson | No scripting |

Several different consensus algorithms of Blockchain are compared in Table VII each of them has its strengths and shortcomings. Table 7 shows a comparative analysis of some essential properties of the Blockchain.

1)ENERGY CONSUMPTION: A massive amount of energy is consumed while mining the block of a Blockchain network (BCN) using proof of work (POW) because of high computation. In proof of stake (POS) and delegated proof of work (DPOS) minimum energy is required for mining as the search area is restricted.

2)DATA MODEL: A data model is a transaction that emphasizes assets. It is required by all the originations to spin the network and exchange the assists.

3)APPLICATION: Ethereum and its derivatives such as Hyperchain, Monaxs, and Parity are used by some ledgers as they let the user write arbitrarily.

The latency of the Blockchain network (BCN) depends upon the number of patients in the Blockchain network (BCN). There is a direct relationship between the latency and the number of patients because if there is an increase in the number of patients in the Blockchain Network (BCN) then the latency will also increase. In decentralized-based storage like Blockchain, latency can be maintained optimal for all the cases which are not achievable in peer-to-peer(P2P) storage. In a worst-case scenario when the number of patients is minimum (100 Patients) a Blockchain-based decentralized storage system can achieve a minimum of 35.64% accuracy whereas in the best case scenario when the number of patients is maximum (500 Patients). In a Blockchain network (BCN) decentralized-based storage can achieve 20% of minimum latency as compared to peer-to-peer (P2P) storage. In the proposed system a Blockchain-based decentralized database is used as it can achieve optimal latency in all the cases when the number of patients is maximum and minimum which is not achievable in centralized storage as shown in Figure 12.

**FIGURE 16. Comparison of Latency**

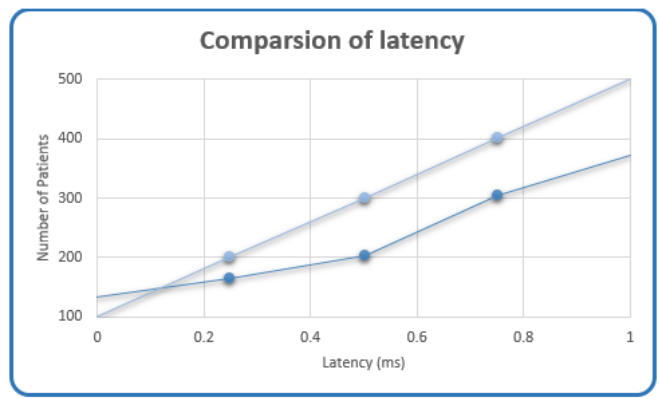

The approach used in the research was 15.51% better precision than the existing approaches. The proposed approach can be used as a base for future research and can be sued to address the emerging problems due to the integration of Blockchain technology with the current systems to anticipate a few ongoing sicknesses that influence inward organs. The proposed approach provides more precise results when the information is gathered from solid or good sources.

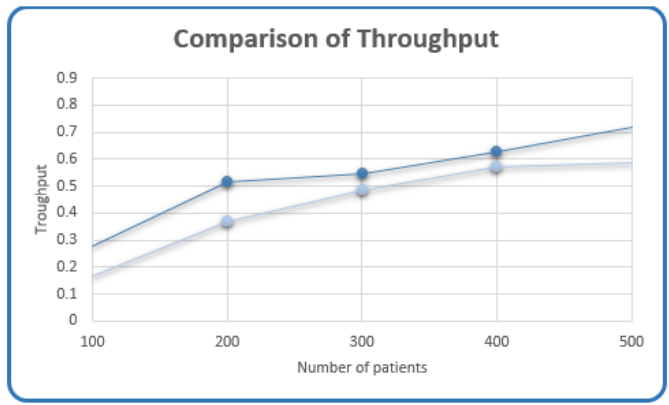

**FIGURE 17. Comparison of Throughput**

## V. CONCLUSION

Blockchain assumes a crucial part in getting along medical services information electronically with an extraordinary degree of protection. This paper specifies, that the significance of utilizing the Blockchain in medical care records is considered and the issue has been all around tended by putting away the clinical information in a Blockchain network. Another significance of this article is expected in the medical services industry is ideal forecast. Subsequently, a metaheuristic approach is utilized to predict coronary illness from the information put away in the Blockchain. The proposed SCA-WKNN accomplishes the most extreme exactness and accuracy in all attributes as shown in the graphs above if used in contrast with the customary calculations. When the sine-cosine weighted k-nearest neighbor's algorithm is used in combination with a weighted k-nearest neighbor and other machine learning algorithms gives the ideal task of the class mark. The proposed Blockchain-based network capacity has maximum throughput and precision when contrasted and unified capacity.